\documentclass[12pt]{article}
\usepackage{xcolor}

\usepackage[a4paper, top=1in, bottom=1in, left=0.85in, right=0.85in]{geometry}

\usepackage{setspace}
\usepackage[T1]{fontenc}
\usepackage[utf8]{inputenc}
\usepackage{newtxtext,newtxmath} 
\usepackage{microtype}

\usepackage{float}
\usepackage{graphicx}
\usepackage{booktabs}
\usepackage{siunitx}
\usepackage{algorithm}
\usepackage{algorithmicx}
\usepackage{algpseudocode}

\usepackage{caption}
\usepackage{pdflscape}

\usepackage{xcolor}
\usepackage{adjustbox}
\usepackage{enumitem}
\usepackage{authblk}
\usepackage{url}
\usepackage{cite}

\newcommand{\code}[1]{\texttt{\hspace{1.5pt}#1}}

\newcommand{\MolFromSmiles}{\code{Chem.\allowbreak MolFromSmiles}}

\title{\textbf{TSSR: Two-Stage Swap-Reward–Driven Reinforcement Learning for Character-Level SMILES Generation}}
\author[1]{Jacob Ede Levine}
\author[2,*]{Yun Lyan Luo}
\author[1,*]{Sai Chandra Kosaraju}
\affil[1]{Department of Computer Science, California State Polytechnic University, Pomona, CA, USA}
\affil[2]{Department of Biotechnology and Pharmaceutical Sciences, Western University of Health Sciences, Pomona, CA, USA}
\affil[*]{Corresponding author: luoy@westernu.edu, skosaraju@cpp.edu}
\begin{document}
\date{}
\maketitle
\begin{abstract}
The design of reliable, valid, and diverse molecules is fundamental to modern drug discovery, as improved molecular generation supports efficient exploration of the chemical space for potential drug candidates and reduces the cost of early design efforts. Despite these needs, current chemical language models that generate molecules as SMILES strings are vulnerable to compounding token errors: many samples are unparseable or chemically implausible, and hard constraints meant to prevent failure can restrict exploration. To address this gap, we introduce TSSR, a Two-Stage, Swap-Reward-driven reinforcement learning (RL) framework for character-level SMILES generation. Stage one rewards local token swaps that repair syntax, promoting transitions from invalid to parseable strings. Stage two provides chemistry-aware feedback from RDKit diagnostics, rewarding reductions in valence, aromaticity, and connectivity issues. The reward decomposes into interpretable terms (swap efficiency, error reduction, distance to validity), is model agnostic, and requires no task-specific labels or hand-crafted grammars. We evaluated TSSR on the MOSES benchmark using a GRU policy trained with PPO in both pure RL (P-RL) from random initialization and fine-tuning RL (F-RL) starting from a pretrained chemical language model, assessing 10{,}000 generated SMILES per run. In P-RL, TSSR significantly improves syntactic validity, chemical validity, and novelty. In F-RL, TSSR preserves drug-likeness and synthesizability while increasing validity and novelty. Token-level analysis shows that syntax edits and chemistry fixes act jointly to reduce RDKit detected errors. TSSR converts a sparse terminal objective into a denser and more interpretable reward, improving both syntactic and chemical quality without reducing diversity. The TSSR framework is dataset-agnostic and can be adapted to various reinforcement learning approaches for de novo drug design.




\end{abstract}
\section{Introduction}
\label{sec1}
De novo molecular design, the computational generation of novel compounds with desired properties, has emerged as a promising paradigm in modern drug discovery \cite{schneider_rethinking_2020, Unlu2025DrugGEN, Loeffler2024REINVENT4, Kadurin2017Cornucopia, blaschke2017applicationgenerativeautoencodernovo, makhzani2016adversarialautoencoders, Prykhodko2019LatentVAEGAN, gulrajani2017improvedtrainingwassersteingans, grisonibidirectional}. Unlike virtual screening, which searches existing libraries, de novo design generates molecules from scratch, offering a strategy to explore the unknown chemical space of potential drug-like compounds. For example, the Enamine REAL Space enumerates approximately 76.9 billion make-on-demand molecules, representing the largest collection of synthetically accessible compounds to date. However, this number represents only a tiny fraction of the chemical space of small molecules, where the count of possible structures with up to 30 heavy atoms is estimated to exceed $10^{60}$. The ability to efficiently generate novel, unique, and chemically valid molecules is therefore a critical challenge in the search for new therapeutics.

Deep learning has become an essential tool in modern de novo molecular design, enabling rapid screening, molecular property prediction, and structure-guided design across various molecular formats~\cite{ADELUSI2022100880,Vamathevan_9,Schneider_2017,corso2022diffdock}. For instance, graph-based approaches demonstrate that neural networks operating directly on molecular graphs can capture atomic connectivity, substructure patterns, and other essential chemical features, supporting a wide range of tasks in computational generation of novel compounds~\cite{Amara2023, Zhao2024,Wang2025}.
Geometric deep learning methods incorporate 3D spatial information to better represent molecular interactions, and structure-based models such as deep docking and learned binding-pose predictors help accelerate virtual screening within de novo molecular design and modern drug discovery~\cite{Isert2023,Hop2018}.
These advances highlight the power of deep learning to represent molecular complexity across multiple modalities. However, these methods often require curated structural data, computationally intensive geometric preprocessing, or large labeled datasets that may not be available for many therapeutic targets~\cite{Askr2023,Lee2025,Li2024,Buttenschoen2023}.

SMILES representations provide a lightweight and scalable way to encode molecules as simple text strings, allowing models to process chemical structures efficiently without relying on curated protein structures, 3D conformers, or other computationally intensive geometric inputs. Their linear format enables fast preprocessing and straightforward compatibility with modern sequence-based architectures, making them well suited for large datasets and high-throughput generative workflows~\cite{Weininger1988SMILESAC,korshunova_generative_2022,segler_generating_2018,popova2019molecularrnngeneratingrealisticmolecular,brown_guacamol_2019}. With SMILES as a sequence representation, chemical language models (CLMs) treat SMILES analogously to natural language, learning chemical syntax, functional group patterns, and scaffold regularities directly from large molecular corpora. By capturing both local and long-range dependencies, CLMs offer an efficient route to explore the chemical space and propose diverse drug-like structures, particularly in early discovery scenarios where detailed structural information may be scarce or unavailable~\cite{GomezBombarelli2018,Chithrananda2020ChemBERTa,Schwaller2021MolTransformer}.

However, molecule generation in SMILES form is inherently fragile because each token \(s_t\) depends on the full prefix \(s_{1:t-1}\). An early token error can cascade through the entire sequence, making long or structurally complex molecules especially vulnerable to failure~\cite{Flam_Shepherd_2022}. Consequently, models often output invalid syntax or misordered tokens that produce unparseable or chemically implausible molecules, particularly when sampling from pretrained models without additional constraints~\cite{pang_deep_2024,GomezBombarelli2018ACS}. To address these issues, many methods impose constraints such as limiting sequence length or using context-aware token selection. Alternative string formats such as DeepSMILES and SELFIES were developed to guarantee syntactic validity~\cite{deepsmiles,selfies2020,selfies2023}, and fragment-based representations including Group SELFIES, SAFE, and fragSMILES restrict generation to chemically meaningful units~\cite{groupselfies2023,safe2024,fragSMILES2025}. Although reducing error rates, these strategies also bias models toward safer and more predictable patterns, which restricts open ended exploration of novel chemistry~\cite{Rajan_Zielesny_Steinbeck_2021,frey2022fastflowsflowbasedmodelsmolecular,nigam2021janusparalleltemperedgenetic,nigam2020augmentinggeneticalgorithmsdeep,odyssey,beautiful}. It has been shown that strict validity enforcement narrow the exploration of chemical space, as allowing the model to encounter some near-invalid or imperfect sequences can help preserve diversity during generation~\cite{Skinnider2024InvalidSMILES}. These challenges highlight the need for a flexible framework that provides dense, chemically informed feedback during generation.

Several prior approaches, including the REINVENT family, adapt SMILES-based generators using REINFORCE-style policy gradients with sparse terminal rewards such as validity checks or thresholded property scores, typically combined with explicit regularization toward a pretrained prior~\cite{olivecrona_molecular_2017,doi:10.1021/acs.jcim.0c00915,Loeffler2024REINVENT4,Thomas2022-zg,doi:10.1021/acs.jcim.5c02053}. Although these methods can produce useful results, sparse end-point feedback makes it difficult to assign credit to early token decisions and often requires strong priors or curriculum heuristics to preserve chemically valid behavior~\cite{doi:10.1126/sciadv.aap7885,Thomas2025-wo}. Many existing pipelines further depend on pretrained models or constrained action spaces to maintain training stability~\cite{olivecrona_molecular_2017,Loeffler2024REINVENT4}. Together with the inherent fragility of sequential SMILES generation, these challenges underscore the need for training strategies that provide richer and chemically informed guidance throughout the generation process rather than relying solely on sparse terminal signals.

In this work, we propose TSSR (Two-Stage Swap-Reward), a reinforcement learning framework designed to address long-standing challenges of syntactic validity, chemical plausibility, and novelty in de novo molecular generation. TSSR introduces a two-stage reward structure that first enforces syntactic correctness through localized token-level repairs, and then improves chemical plausibility by applying graded penalties for structural inconsistencies, as illustrated in Figure~\ref{fig:algorithm_application}. We view molecule construction as a sequential decision-making problem in which each token depends on the preceding context. This formulation is well captured by a Markov Decision Process, allowing reinforcement learning to optimize the generative model using terminal rewards derived from syntactic repair and chemistry diagnostics. By contrast with previous approaches, TSSR provides richer terminal feedback and does not require a specialized vocabulary or a pretraining step, making the framework model- and dataset agnostic. It can be used to train a chemical language model entirely from scratch or to fine-tune an existing model purely through reinforcement learning using chemically informed reward signals.

As the generative backbone, we adopt a character-level recurrent neural network, specifically a Gated Recurrent Unit (GRU), which naturally models the left-to-right token dependencies present in SMILES. GRU-based chemical language models have been widely used for molecular generation, yet they typically rely on maximum-likelihood pretraining on large datasets to achieve acceptable validity. TSSR relaxes this requirement by providing a chemically grounded terminal reward that enables a GRU policy to learn valid and diverse molecules directly through reinforcement learning. To demonstrate the flexibility of this framework, we evaluate TSSR under two complementary training regimes. In the pure reinforcement learning setting (P-RL), the GRU is trained entirely from random initialization using only TSSR feedback. In the fine-tuning setting (F-RL), the GRU is initialized from a pretrained SMILES language model and then optimized with TSSR. Together, these regimes show that TSSR supports both training from scratch and reward-guided refinement of pretrained models, enabling broad applicability across molecular design workflows.

The two-stage reward formulation enables a randomly initialized GRU to learn SMILES syntax and generate chemically meaningful structures without requiring any supervised pretraining. When applied to fine-tuning, TSSR further improves the generation quality of an already competent language model by correcting residual syntactic and chemical errors. As presented in the Results section, P-RL training with TSSR yields a substantial increase in syntactic validity, roughly doubles the proportion of chemically valid molecules compared to the untrained baseline, and produces a large number of novel structures not present in the training set. In the F-RL setting, where the pretrained model already achieves high baseline validity, TSSR still provides measurable gains by adding additional valid and unique molecules and enhancing overall chemical correctness. The simplicity, flexibility, and model-agnostic nature of TSSR make it straightforward to incorporate into a wide range of sequence-generation pipelines, and the full methodological details are provided in the Methods section.

\section{Results}\label{sec2}

TSSR is applied to improve both the validity and novelty of molecular generation on the MOSES dataset. The first stage of the reward encourages syntactic correctness by rewarding successful repairs that convert invalid SMILES into valid representations. The second stage refines these candidates by rewarding reductions in chemical errors among sequences that remain valid after syntax correction. The effectiveness of this two-stage design was assessed through standard reinforcement learning metrics, such as peak episode reward and throughput, as well as molecular generation metrics, including syntactic validity, chemical validity, novelty, quantitative estimate of drug-likeness (QED), Synthetic accessibilty (SA), diversity, and uniqueness. These evaluations demonstrate the quantitative impact of TSSR on molecular generation quality.

\subsection{Dataset and preprocessing}

The experiments are conducted using the MOSES benchmark dataset, which is curated from the ZINC database and derived from the ZINC Clean Leads collection following standard filtering and preprocessing protocols~\cite{polykovskiy2020molecularsetsmosesbenchmarking}. The dataset comprises approximately 1.9 million molecular SMILES strings, from which 1.6 million molecules are used for the training set and 0.3 million molecules for the test set. This large and chemically diverse dataset provides a comprehensive benchmark for evaluating our proposed $TSSR$ reward framework.  

To prepare the data for training, each SMILES strings are tokenized into a vocabulary set \(V\), where each token \(v \in V\) represents a fundamental syntactic or chemical unit. The vocabulary included atom tokens such as \(C, N, O, S, F, Cl, Br\), bond tokens such as \(c, n, -, =, \#\), ring indices from \(0\) to \(9\), and branching symbols represented by the characters \((\) and \()\). Additionally, special tokens are included such as \texttt{[BOS]} (Beginning of String), \texttt{[EOS]} (End of String), and \texttt{[PAD]} (Padding), which are essential for initializing the generation process, signaling sequence termination, and enabling efficient batch processing during training.  

The global prior frequency distribution of each token in \(V\), excluding the special tokens, are computed from the complete MOSES dataset. These priors were consistently applied during the training of our TSSR framework for both P-RL and F-RL. Incorporating prior token frequencies ensured that the model remained aligned with the underlying chemical distribution of real molecules while still allowing exploration and optimization guided by the reward function.

\subsection{Syntax Repair and Chemistry Fixes}
\begin{figure}[!htbp]
    \begin{adjustbox}{max width=\textwidth}
        \includegraphics[scale=0.4]{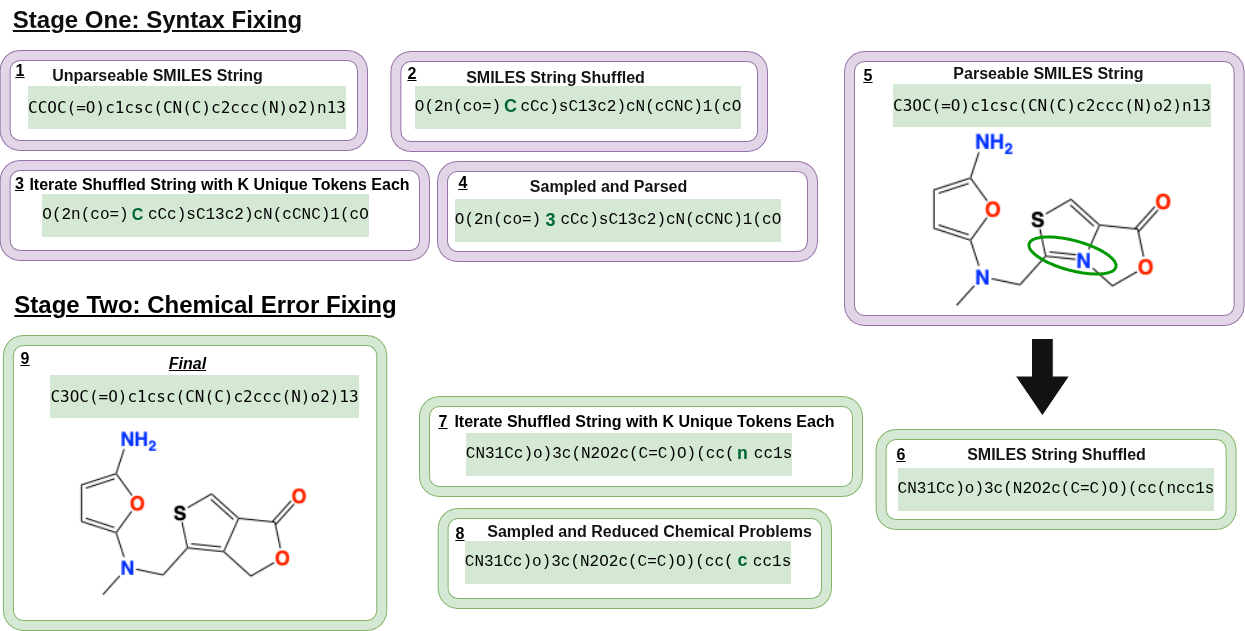}
    \end{adjustbox}
    \caption{Example a Two-Stage, Swap-Reward-driven (TSSR) reinforcement learning (RL) framework for character-level SMILES
generation.}
   \label{fig:algorithm_application}
   \vspace{1cm}
\end{figure}

In \textbf{Figure~\ref{fig:algorithm_application}}, the TSSR is shown repairing an unparseable SMILES strings. The process begins by shuffling the original sequence to prevent positional bias toward fixing tokens that appear earlier in the string. TSSR then iterates through each token position and samples \(K\) candidate tokens based on the global frequency priors. For each sampled token, it replaces the current token and checks whether the modified sequence can be parsed. In this example, the algorithm eventually reaches the first \(C\) token and replaces it with \(c\), allowing the SMILES strings to become syntactically valid. Once a valid string is obtained, the sequence is unshuffled and examined for chemical problems. A single chemistry problem is detected, triggering Stage Two of the TSSR process. The sequence is shuffled again and the same procedure is repeated. For each token position, \(K\) new candidate tokens are sampled and tested for replacements that reduce the number of chemical issues. In this case, replacing token \(n\) with \(c\) removes the remaining chemistry problem, and the algorithm terminates with a fully syntactically and chemically valid molecule.

The two stages of the TSSR framework were evaluated at the token level by quantifying the swaps per molecule (swap count) during Stage One, the fix rate during Stage Two, the chemical error reduction following Stage Two, and the length of the generated SMILES sequences. The swap count was defined as the number of tokens replaced during syntax repair in Stage One, normalized by the length of the generated SMILES sequence. It is defined as:

\begin{equation}
    \text{Swap Count} = \frac{1}{N}\sum_{j=1}^{N} N_{\text{swap}}^{(j)}.
\end{equation}

where, \(N\) is the total number of generated sequences in the run, and \(N_{\text{swap}}^{(j)}\) is the number of token swaps applied to sequence \(j\) during Stage One (zero if no swaps were needed). In the P-RL setting, the swap count increased from \(0.69\) swaps per molecule at \(T_0\) to \(15.51\) swaps per molecule at \(T_{\hat{R}_{\max}}\). Although swap count is generally expected to decrease as the model learns to generate syntactically correct sequences, the observed increase is explained by the complexity of molecules increasing leading to more difficult fixes occurring, hence taking more swaps. In the F-RL setting, the swap count  increased from \(39.00\) at \(T_0\) to \(40.09\) at \(T_{\hat{R}_{\max}}\). This increase indicates that the TSSR framework is increasing the complexity of molecules generated.

In Stage Two, TSSR performs chemical level corrections by modifying tokens that cause structural errors, such as impossible valence states, incorrect ring closures, or chemically implausible bonding patterns. The fix rate was defined as the number of chemical level token corrections performed after Stage One, normalized by sequence length. It is defined as:

\begin{equation}
    \text{Fix Rate} = \frac{N_{\text{fixed}}}{N}.
\end{equation}

where, \(N\) is the total number of generated molecules in the run, and \(N_{\text{fixed}}\) is the number of molecules that were originally syntactically invalid but were successfully repaired to parseable SMILES by Stage One. Molecules that were already syntactically valid without repair are not counted in \(N_{\text{fixed}}\). In the P-RL setting, the fix rate increased from \(0.06\) to \(0.26\) between \(T_0\) and \(T_{\hat{R}_{\max}}\). In the F-RL setting, the fix rate increased  from \(0.26\) at \(T_0\) to \(0.27\) at \(T_{\hat{R}_{\max}}\), suggesting more generated molecules are able to be fixed. This increase is consistent with the rise in average SMILES quality, as the distribution of SMILES strings generated gets closer to being valid, the more the algorithm is able to fix. These results indicate that the two-stage reward mechanism operates in a synchronized manner. Syntax-level improvements in Stage One are complemented by targeted chemical corrections in Stage Two, resulting in an overall reduction in structural errors during molecular generation.

Chemical error reduction is quantified after Stage Two to directly measure the effectiveness of TSSR in resolving chemical issues. It is defined as:

\begin{equation}
    \text{Chemical Error Reduction}=\frac{1}{N}\sum_{j=1}^{N}\left(E_{j}^{(T_0)}-E_{j}^{(T_{\hat{R}_{\max}})}\right)
\end{equation}

where \(E_{j}^{(T_0)}\) is the number of RDKit detected chemistry errors for the \(j\)-th generated SMILES at checkpoint \(t \in \{T_0, T_{\hat{R}_{\max}}\}\), and \(N=10{,}000\) is the number of sampled sequences. By this definition, \(\Delta\mathrm{ChemErr}>0\) indicates improvement (fewer errors after training). Units are errors per sequence. In the P-RL setting, chemical error reduction improved significantly, increasing from \(0.05\) at \(T_0\) to \(0.67\) at \(T_{\hat{R}_{\max}}\). This improvement confirms that TSSR is highly effective in reducing chemical errors during molecular generation, as more are becoming syntactically valid the chemical errors being fixed is rising. In the F-RL setting, the overall chemical error reduction remained stable  at \(0.29\). The total number of valid SMILES increased, indicating that Stage Two chemical fixes remain effective in improving the validity of generated molecules.
\textbf{Tables~\ref{tab:P-RL-token-repair-new}} and \textbf{~\ref{tab:F-RL-token-repair-new}} summarize the swap count, fix rate, and chemical error reduction results for the P-RL and F-RL settings, respectively.

\begin{table*}[!t]
  \footnotesize
  \caption{Token-level repair analysis - P-RL}
  \label{tab:P-RL-token-repair-new}
  \begin{adjustbox}{center, max width=\textwidth}
    \begin{tabular}{@{}l l l l@{}}
        \toprule
        \textbf{Run} & \textbf{Swaps} & \textbf{Fixes} & \textbf{Chem-Err} \\
        \midrule
        1 & 0.69 $\to$ 15.51  & 0.06 $\to$ 0.26 & 0.04 $\to$ 0.34 \\
        2 & 0.73 $\to$ 6.60   & 0.06 $\to$ 0.25 & 0.04 $\to$ 0.32 \\
        3 & 0.79 $\to$ 12.57  & 0.07 $\to$ 0.32 & 0.05 $\to$ 0.59 \\
        4 & 0.80 $\to$ 19.03  & 0.07 $\to$ 0.39 & 0.05 $\to$ 1.15 \\
        5 & 0.78 $\to$ 20.52  & 0.07 $\to$ 0.22 & 0.05 $\to$ 0.96 \\
        \midrule
        \textbf{Mean $\pm$ SD} & 0.76 $\pm$ 0.04 $\to$ 14.85 $\pm$ 4.97 & 0.06 $\pm$ 0.00 $\to$ 0.29 $\pm$ 0.06 & 0.05 $\pm$ 0.00 $\to$ 0.67 $\pm$ 0.33 \\
        \addlinespace
        \textbf{Average SMILES Length} & \multicolumn{3}{c}{26.09 $\to$ 11.90} \\
        \bottomrule
    \end{tabular}
  \end{adjustbox}
  \vspace{1cm}
\end{table*}

\begin{table*}[!t]
  \footnotesize
  \caption{Token-level repair analysis - F-RL}
  \label{tab:F-RL-token-repair-new}
  \begin{adjustbox}{center, max width=\textwidth}
    \begin{tabular}{@{}l l l l@{}}
        \toprule
        \textbf{Run} & \textbf{Swaps} & \textbf{Fixes} & \textbf{Chem-Err} \\
        \midrule
        1 & 39.00 $\to$ 40.09   & 0.26 $\to$ 0.27 & 0.27 $\to$ 0.29 \\
        2 & 18.02 $\to$ 16.90   & 0.14 $\to$ 0.13 & 0.10 $\to$ 0.09 \\
        3 & 59.73 $\to$ 55.96   & 0.28 $\to$ 0.27 & 0.32 $\to$ 0.32 \\
        4 & 58.61 $\to$ 57.02   & 0.40 $\to$ 0.39 & 0.30 $\to$ 0.30 \\
        5 & 103.31 $\to$ 101.91 & 0.34 $\to$ 0.32 & 0.47 $\to$ 0.46 \\
        \midrule
        \textbf{Mean $\pm$ SD} & 55.73 $\pm$ 28.25 $\to$ 54.38 $\pm$ 27.85 & 0.28 $\pm$ 0.09 $\to$ 0.28 $\pm$ 0.09 & 0.29 $\pm$ 0.12 $\to$ 0.29 $\pm$ 0.12 \\
        \addlinespace
        \textbf{Average SMILES Length} & \multicolumn{3}{c}{46.88 $\to$ 46.84} \\
        \bottomrule
    \end{tabular}
  \end{adjustbox}
  \vspace{1cm}
\end{table*}

\begin{figure}[!htbp]
    \begin{adjustbox}{center}
        \includegraphics[scale=0.27]{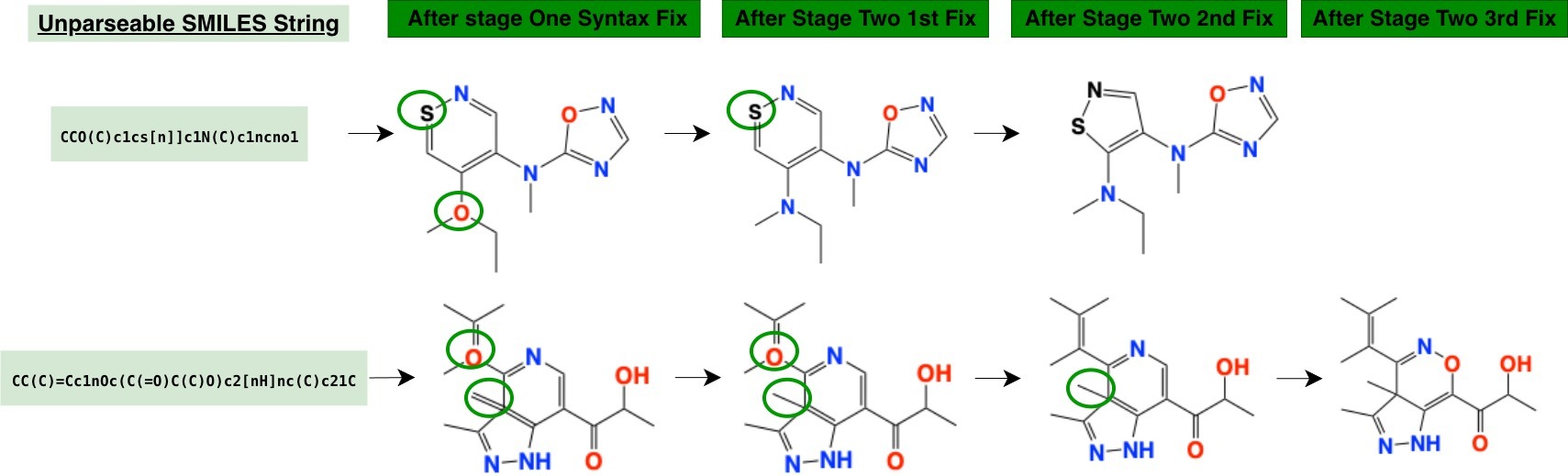}
    \end{adjustbox}  
    \caption{Examples of TSSR Stage Two fixes: Invalid SMILES to Chemically valid with upto 3 Fixes Each}
    \label{fig:invalid_to_perfect}
    \vspace{1cm}
\end{figure}

The number of chemical fixes in Stage Two ranges from one to five, with one to two steps being the most common scenario. In \textbf{Figure~\ref{fig:invalid_to_perfect}} two cases are shown where TSSR repairs SMILES strings in two steps. Each example begins with an unparseable string. After Stage One a local substitution lets RDKit parse the string, which exposes two chemistry problems. Stage Two then proceeds in two passes within one shuffle. In each pass the method samples \(K\) candidate tokens at each position and accepts a change only when the number of detected problems decreases. The first pass reduces the count from two to one, and the second pass removes the final issue. The accepted edits correct local valence or ring closure mistakes, so the RDKit diagnostics fall from two to zero by the end of the sequence of fixes. 

\subsection{Molecular generation using TSSR}
\label{ssec:sequence-quality}

The qualitative performance of TSSR is evaluated using the widely adopted molecular generation metrics of \emph{validity} and \emph{novelty}. To assess the improvements achieved through TSSR training, The validity and novelty of molecules generated by the GRU-based model are compared before training (initial model) and after training with the proposed TSSR reward framework. TSSR uses four hyper-parameters, the first deciding the number of tokens sampled from priors, and the final three determining the weights of the reward terms. Our hyper-parameter tuning is discussed further in (Sec.~\ref{ssec:training}).

For each experiment, \(10{,}000\) SMILES strings were generated to evaluate both metrics. During generation, the first token was fixed as \texttt{[BOS]} (Beginning of String), while subsequent tokens were sampled using softmax multinomial sampling. This sampling approach ensured that the model does not always select the most probable token, thus preserving diversity, while still assigning higher probabilities to more likely tokens.

\paragraph{Validity}
Validity is defined as the proportion of generated SMILES strings that could be successfully parsed into chemically valid molecules using the cheminformatics toolkit \texttt{RDKit}~\cite{greg_landrum}:

\begin{equation}
    \text{Validity} = \frac{N_{\text{valid}}}{N_{\text{gen}}}
\end{equation}

where \(N_{\text{valid}}\) denotes the number of valid molecules among the generated samples, and \(N_{\text{gen}}\) is the number of generated molecules.

In the P-RL setting, we measured the effect of TSSR by comparing model performance at two points. \(T_0\) represents the initial model before training, with randomly initialized parameters. \(T_{\hat{R}_{\max}}\) is the checkpoint at which the model achieved its peak reward \(\hat{R}_{\max}\). TSSR increased the syntactic validity of generated molecules by an average of \(470\%\) (from \(6.14\%\) at \(T_0\) to \(35.03\%\) at \(T_{\hat{R}_{\max}}\)) \textbf{(Table~\ref{tab:P-RL-quality})}. TSSR increased the chemical validity of generated molecules by an average of \(101.47\%\) (from \(4.77\%\) at \(T_0\) to \(9.61\%\) at \(T_{\hat{R}_{\max}}\)).

Next, TSSR is evaluated in the F-RL setting. Here, \(T_0\) represents the pretrained model before fine-tuning, and \(T_{\hat{R}_{\max}}\) is again the checkpoint corresponding to peak reward. In this case, TSSR increased the validity, on average, by \(0.83\%\) (from \(35.36\%\) at \(T_0\) to \(36.36\%\) at \(T_{\hat{R}_{\max}}\)). This improvement corresponds to 83 additional valid SMILES molecules on average generated after fine-tuning (\textbf{Table~\ref{tab:F-RL-quality}}). Chemical Validity on the other hand increases  on average (from \(19.07\%\) at \(T_0\) to \(19.20\%\) at \(T_{\hat{R}_{\max}}\)) resulting in a gain of 13 molecules on average. Together, these results demonstrate that TSSR consistently improves the performance of both models syntactic validity trained from scratch and pretrained chemical language models.

\paragraph{Novelty}
The novelty of generated SMILES is defined as:

\begin{equation}
   \text{Novelty} = \frac{N_{\text{novel}}}{N_{\text{valid}}} 
\end{equation}

where \(N_{\text{novel}}\) is the number of valid molecules that were not present in the training dataset. Novelty measures the proportion of new molecules among all valid ones and is a key indicator of a model's generative capability.

In the P-RL setting, TSSR improved novelty by \(228.72\%\), increasing from \(17.83\%\) at \(T_0\) to \(58.64\%\) at \(T_{\hat{R}_{\max}}\). In the F-RL setting, the pretrained model achieved a novelty of \(99.65\%\) at \(T_0\), compared to \(99.64\%\) at \(T_{\hat{R}_{\max}}\). Although this represents a slight decrease in percentage, the total number of novel molecules increased from 3529 to 3611 on average, corresponding to 82 additional novel molecules generated after fine-tuning. \textbf{Tables~\ref{tab:P-RL-quality}} and\textbf{~\ref{tab:F-RL-quality}} summarize the results of F-RL and P-RL respectively.

These findings demonstrate that the two-stage reward mechanism in TSSR not only improves syntactic correctness and overall validity but also enhances the model's ability to generate novel and chemically meaningful molecules. By providing chemically aware feedback during training, TSSR drives the generation process toward both higher-quality and more diverse molecular outputs.

\begin{table*}[!t]
  \footnotesize
  \caption{Molecular quality metrics - P-RL}
  \label{tab:P-RL-quality}
  \begin{adjustbox}{center, max width=\textwidth}
      \begin{tabular}{@{}l l l l l l@{}}
        \toprule
        \textbf{Run} & \textbf{Syntactic Val (\%)} & \textbf{Novel (\%)} & \textbf{Novel (count)} & \textbf{Chem Val (\%)} & \textbf{Chem Val (count)} \\
        \midrule
        1 & 6.37 $\to$ 75.97* & 17.74 $\to$ 79.37* & 113 $\to$ 6,030 & 4.97 $\to$ 12.04* & 497 $\to$ 1,204 \\
        2 & 5.87 $\to$ 32.02* & 17.55 $\to$ 76.83* & 103 $\to$ 2,460 & 4.57 $\to$ 6.61* & 457 $\to$ 661 \\
        3 & 6.74 $\to$ 23.27* & 16.62 $\to$ 59.22* & 112 $\to$ 1,378 & 5.32 $\to$ 7.98* & 532 $\to$ 798 \\
        4 & 5.72 $\to$ 22.09* & 18.53 $\to$ 34.18* & 106 $\to$ 755 & 4.35 $\to$ 12.43* & 435 $\to$ 1,243 \\
        5 & 6.04 $\to$ 21.79* & 18.71 $\to$ 43.60* & 113 $\to$ 950 & 4.62 $\to$ 8.99* & 462 $\to$ 899 \\
        \midrule
        \textbf{Mean$\pm$SD} 
          & 6.15 $\pm$ 0.41 $\to$ 
          & 17.83 $\pm$ 0.84 $\to$ 
          & 109 $\pm$ 5 $\to$ 
          & 4.77 $\pm$ 0.38 $\to$ 
          & 477 $\pm$ 38 $\to$  \\
          & 35.03 $\pm$ 23.27
          & 58.64 $\pm$ 19.91
          & 2,315 $\pm$ 2,179
          & 9.61 $\pm$ 2.54
          & 961 $\pm$ 254 \\
        \bottomrule
      \end{tabular}
  \end{adjustbox}
  * denotes a statistically significant difference by a 95\% confidence p-test.+
  \vspace{1cm}
\end{table*}

\begin{table*}[!t]
  \footnotesize
  \caption{Molecular quality metrics - F-RL}
  \label{tab:F-RL-quality}
  \begin{adjustbox}{center, max width=\textwidth}
        \begin{tabular}{@{}l l l l l l@{}}
        \toprule
        \textbf{Run} & \textbf{Syntactic Val (\%)} & \textbf{Novel (\%)} & \textbf{Novel (count)} & \textbf{Chem Val (\%)} & \textbf{Chem Val (count)} \\
        \midrule
        1 & 37.00 $\to$ 38.36 & 99.92 $\to$ 99.92 & 3,697 $\to$ 3,833 & 10.33 $\to$ 10.11 & 1,033 $\to$ 1,011 \\
        2 & 30.07 $\to$ 30.95 & 99.87 $\to$ 99.81 & 3,003 $\to$ 3,089 & 13.41 $\to$ 14.22 & 1,341 $\to$ 1,422 \\
        3 & 12.77 $\to$ 13.59 & 99.92 $\to$ 100.00 & 1,276 $\to$ 1,359 & 3.71 $\to$ 3.88 & 371 $\to$ 388 \\
        4 & 15.46 $\to$ 15.98 & 99.87 $\to$ 99.81 & 1,544 $\to$ 1,595 & 6.45 $\to$ 6.42 & 645 $\to$ 642 \\
        5 & 82.36 $\to$ 82.94 & 98.65 $\to$ 98.64 & 8,125 $\to$ 8,181 & 61.43 $\to$ 61.39 & 6,143 $\to$ 6,139 \\
        \midrule
        \textbf{Mean$\pm$SD} 
          & 35.53 $\pm$ 28.04 $\to$ 
          & 99.65 $\pm$ 0.56 $\to$ 
          & 3,529 $\pm$ 2,759 $\to$ 
          & 19.07 $\pm$ 23.97 $\to$ 
          & 1,907 $\pm$ 2,397 $\to$ \\
        & 36.36 $\pm$ 28.00
          & 99.64 $\pm$ 0.56
          & 3,611 $\pm$ 2,754
          & 19.20 $\pm$ 23.90
          &  1,920 $\pm$ 2,390 \\
        \bottomrule
      \end{tabular}
  \end{adjustbox}
  \vspace{1cm}
\end{table*}

\subsection{Drug-likeness, synthesizability, diversity, and uniqueness}
While the focus of the current study is chemical validity and novelty, three quality indicators were examined that are commonly used in de novo drug design. Quantitative Estimate of Drug-likeness (QED) summarizes how druglike a molecule is on a 0 to 1 scale, where higher is better. The synthetic accessibility (SA) score estimates ease of synthesis, with a lower score indicating greater ease. Uniqueness measures the fraction of non-duplicate molecules in the generated set. Diversity is a measure of the similarity between molecules, further defined in~\ref{Metrics and Evaluation}, defines how similar on average the molecules in the generation are to eachother, with a higher score indicating higher diversity. Scaffold count is the amount of unique scaffolding the molecules posses as compared to the original dataset.

 In P-RL tests, however, starting from scratch produced a different profile (\textbf{Table~\ref{tab:P-RL-qed-sa}}). Mean QED decreased from \(0.374\) to \(0.365\), although the count with QED \(\ge 0.6\) increased modestly from \(10\)  to \(90\). SA increased from \(4.055\) to \(4.371\) on average, which suggests a slight shift toward  harder to synthesize structures on this metric. Diversity decreased slightly from \(0.686\) to \(0.649\) while scaffold count increased from \(14\) to \(83\). Uniqueness again remained at 1.000 for all runs. In F-RL tests, QED remained stable with a small average increase from 0.664 to 0.667 across runs (\textbf{Table~\ref{tab:F-RL-qed-sa}}). The count of molecules with QED \(\ge 0.6\) changed on average, from \(6920\) to \(6890\). SA also stayed essentially unchanged, moving from \(4.779\) to \(4.790\) on average, with a similar spread. Uniqueness was 1.000 before and after in all runs. Diversity on average stayed stable at \(0.674\) and scaffold diversity decreased slightly from \(2,731\) to \(2709\). Together these results show that F-RL preserves drug-likeness and synthesizability while improving validity and novelty. P-RL improves validity and novelty from a blank start, but it also tends to explore simpler motifs that can reduce average QED and raise SA.

\begin{table*}[!t]
  \footnotesize
  \caption{QED, SA, diversity, scaffold, and uniqueness - P-RL}
  \label{tab:P-RL-qed-sa}
  \begin{adjustbox}{center, max width=\textwidth}
    \begin{tabular}{@{}l l l l l l l@{}}
        \toprule
        \textbf{Run} & \textbf{QED mean} & \textbf{QED $\ge$ 0.6 (count)} & \textbf{SA mean} & \textbf{Diversity} & \textbf{Scaffold count} & \textbf{Uniq} \\
        \midrule
        1 & 0.371 $\to$ 0.326 & 0 $\to$ 0 & 4.030 $\to$ 4.705  & 0.680 $\to$ 0.661 & 10 $\to$ 57 & 1.000  \\
        2 & 0.367 $\to$ 0.358 & 0 $\to$ 150 & 4.158 $\to$ 4.276  & 0.694 $\to$ 0.638 & 20 $\to$ 102 & 1.000  \\
        3 & 0.375 $\to$ 0.363 & 0 $\to$ 90 & 4.001 $\to$ 4.509  & 0.683 $\to$ 0.652 & 11 $\to$ 134 & 1.000  \\
        4 & 0.383 $\to$ 0.383 & 0 $\to$ 210 & 4.036 $\to$ 4.333  & 0.681 $\to$ 0.638 & 17 $\to$ 96 & 1.000  \\
        5 & 0.374 $\to$ 0.394 & 7 $\to$ 0 & 4.047 $\to$ 4.031  & 0.691 $\to$ 0.656 & 12 $\to$ 28 & 1.000  \\
        \midrule
        \textbf{Mean$\pm$SD}
          & 0.374 $\pm$ 0.006 $\to$ 
          & 10 $\pm$ 30 $\to$ 
          & 4.055 $\pm$ 0.060 $\to$
          & 0.686 $\pm$ 0.006 $\to$
          & 14 $\pm$ 4 $\to$ \\
          & 0.365 $\pm$ 0.026
          & 90 $\pm$ 90
          & 4.371 $\pm$ 0.253
          & 0.649 $\pm$ 0.010
          & 83 $\pm$ 41 \\
        \bottomrule
    \end{tabular}
  \end{adjustbox}
  \vspace{1cm}
\end{table*}

\begin{table*}[!t]
  \footnotesize
  \caption{QED, SA, diversity, scaffold, and uniqueness - F-RL}
  \label{tab:F-RL-qed-sa}
  \begin{adjustbox}{center, max width=\textwidth}
    \begin{tabular}{@{}l l l l l l l@{}}
        \toprule
        \textbf{Run} & \textbf{QED mean} & \textbf{QED $\ge$ 0.6 (count)} & \textbf{SA mean} & \textbf{Diversity} & \textbf{Scaffold count} & \textbf{Uniq} \\
        \midrule
        1 & 0.594 $\to$ 0.591 & 5590 $\to$ 5460 & 4.913 $\to$ 4.944  & 0.657 $\to$ 0.658 & 1,696 $\to$ 1,673 & 1.000 \\
        2 & 0.680 $\to$ 0.685 & 7270 $\to$ 7270 & 4.655 $\to$ 4.628  & 0.695 $\to$ 0.696 & 3,757 $\to$ 3,765 & 1.000 \\
        3 & 0.640 $\to$ 0.646 & 6250 $\to$ 6360 & 4.983 $\to$ 5.008  & 0.727 $\to$ 0.727 & 1,721 $\to$ 1,690 & 1.000 \\
        4 & 0.636 $\to$ 0.637 & 6370 $\to$ 6240 & 4.855 $\to$ 4.878  & 0.697 $\to$ 0.702 & 2,176 $\to$ 2,210 & 1.000 \\
        5 & 0.773 $\to$ 0.774 & 9120 $\to$ 9130 & 4.491 $\to$ 4.490  & 0.595 $\to$ 0.588 & 4,305 $\to$ 4,206 & 1.000 \\
        \midrule
        \textbf{Mean$\pm$SD}
          & 0.664 $\pm$ 0.068 $\to$ 
          & 6920 $\pm$ 1370 $\to$ 
          & 4.779 $\pm$ 0.202 $\to$ 
          & 0.674 $\pm$ 0.051 $\to$
          & 2,731 $\pm$ 1,218 $\to$ \\
          & 0.667 $\pm$ 0.069
          & 6890 $\pm$ 1410
          & 4.790 $\pm$ 0.221
          & 0.674 $\pm$ 0.054
          & 2,709 $\pm$ 1,195 \\
        \bottomrule
    \end{tabular}
  \end{adjustbox}
  \vspace{1cm}
\end{table*}

\subsection{Computational evaluation of TSSR}
The learning and computational efficiency of the proposed TSSR framework are evaluated using two complementary metrics: the peak episode reward \(\hat{R}_{\max}\) and the throughput \(\tau\). The peak reward \(\hat{R}_{\max}\) measures how effectively the GRU-RNN learns to generate valid and novel SMILES strings based on the feedback provided by TSSR. It is defined as:

\begin{equation}
\hat{R}_{\max} = \max_{e \in \{1, \dots, E\}} \sum_{t=1}^{T} \gamma^{t-1} r_t
\label{eq:rmax}
\end{equation}

where \(r_t\) denotes the reward obtained at step \(t\), \(\gamma = 0.99\) is the discount factor, and \(E\) is the total number of training episodes. A higher \(\hat{R}_{\max}\) indicates that the model has learned the reward landscape and is generating more chemically valid and novel molecules.

Computational efficiency is quantified using throughput \(\tau\), which measures the number of SMILES tokens the GRU-RNN can process within a given time \(T\). Throughput is defined as:

\begin{equation}
\tau = \frac{N_g \times B \times L}{T}
\label{eq:tau}
\end{equation}

where \(N_g\) is the number of gradient updates, \(B = 512\) is the batch size, and \(L \) is the average SMILES sequence length per run. A higher \(\tau\) reflects the model's ability to process a larger number of tokens per unit time.

These two metrics capture distinct but complementary aspects of model performance. While \(\hat{R}_{\max}\) measures \emph{learning efficiency} (how well the model optimizes its policy), \(\tau\) reflects \emph{computational efficiency} (how quickly the model generates molecules). All reported results are averaged over five independent runs to ensure robustness and reproducibility. \textbf{Tables~\ref{tab:P-RL-eff} and~\ref{tab:F-RL-eff}} summarize the results for P-RL and F-RL, respectively, using \(N_g = 9000\) gradient updates. P-RL achieves a peak reward of \(\hat{R}_{\max} = 0.742\) with a throughput of \(\tau = 52\), while F-RL attains \(\hat{R}_{\max} = 0.549\) with a higher throughput of \(\tau = 189\). The higher average throughput observed in F-RL reflects the reduced computational cost due to pre-training, whereas the higher peak reward achieved by P-RL indicates superior learning efficiency, which is also reflected in the improved validity and novelty of the generated molecules.

\begin{table*}[!t]
  \footnotesize
  \caption{Training efficiency - P-RL}
  \label{tab:P-RL-eff}
  \begin{adjustbox}{center, max width=\textwidth}
    \begin{tabular}{@{}l l l l@{}}
      \toprule
      \textbf{Run} & $\hat R_{\max}$ & \textbf{Time (s)} & $\tau$ (kTok/s) \\
      \midrule
      1 & 0.591 & 1042.7 & 87 \\
      2 & 0.749 & 1088.3 & 34 \\
      3 & 0.756 & 1007.5 & 38 \\
      4 & 0.808 & 1055.2 & 42 \\
      5 & 0.805 & 1096.6 & 58 \\
      \midrule
      \textbf{Mean $\pm$ SD} & 0.742 $\pm$ 0.079 & 1058.060 $\pm$ 36.056 & 52 $\pm$ 21 \\
      \bottomrule
    \end{tabular}
  \end{adjustbox}
  \vspace{1cm}
\end{table*}

\begin{table*}[!t]
  \footnotesize
  \caption{Training efficiency - F-RL}
  \label{tab:F-RL-eff}
  \begin{adjustbox}{center, max width=\textwidth}
    \begin{tabular}{@{}l l l l@{}}
      \toprule
      \textbf{Run} & $\hat R_{\max}$ & \textbf{Time (s)} & $\tau$ (kTok/s) \\
      \midrule
      1 & 0.533 & 1112.4 & 183 \\
      2 & 0.571 & 1175.9 & 144 \\
      3 & 0.567 & 1133.3 & 194 \\
      4 & 0.513 & 1188.8 & 161 \\
      5 & 0.562 & 1121.0 & 263 \\
      \midrule
      \textbf{Mean $\pm$ SD} & 0.549 $\pm$ 0.022 & 1146.280 $\pm$ 34.061 & 189 $\pm$ 46 \\
      \bottomrule
    \end{tabular}
  \end{adjustbox}
  \vspace{1cm}
\end{table*}

\section{Discussion}\label{sec:discussion}

This study demonstrates that the TSSR framework can transform the inherently sparse and terminal nature of SMILES generation into structured and interpretable feedback. In Stage One, the agent focuses on local token substitutions that repair syntactic errors, bringing sequences into parseable regions of the chemical grammar. In Stage Two, it favors edits that lower RDKit chemistry diagnostics, guiding molecules toward chemically coherent and realistic structures. Together, these two stages form a progressive learning pathway: first improving syntactic precision, then advancing chemical plausibility.

The effects of this structured reward are evident in the token-level repair statistics reported in Table~\ref{tab:P-RL-token-repair-new} and Table~\ref{tab:F-RL-token-repair-new}. For P-RL, the average number of token swaps increases markedly from $0.76$ to $14.85$ per episode, while the fraction of successfully repaired sequences rises from $0.06$ to $0.29$. At the same time, the average chemical error metric (\textit{Chem-Err}) grows moderately from $0.05$ to $0.67$, reflecting that the model is attempting and often succeeding to modify a larger portion of invalid sequences. The simultaneous rise in swap activity and fix rate shows that the model learns to explore and correct syntax more aggressively, a behavior aligned with Stage One repair dynamics. Notably, the average SMILES length drops from $26.09$ to $11.90$, indicating that shorter, more manageable strings tend to be easier to fix and validate under the TSSR reward.

In contrast, the fine-tuning setup (F-RL) in Table~\ref{tab:F-RL-token-repair-new} shows much smaller changes between initial and final statistics. The number of swaps remains decreases slightly from ($55.73\!\to\!54.38$ on average), and both fix and chemical error rates stay nearly constant at $0.28$ and $0.29$, respectively. This stability arises because the pretrained model already follows SMILES grammar closely. The reward thus provides less new information for Stage One, and the main improvements occur in Stage Two, where small chemistry refinements dominate. The steady SMILES length ($46.88\!\to\!46.84$) further suggests that the pretrained prior maintains consistent sequence structure throughout learning.

The performance gap in peak reward \(\hat{R}_{\max}\) between P-RL and F-RL stems from two factors. First, P-RL uses a higher learning rate (\textbf{Table~\ref{tab:rl-hparams}}), allowing the policy to move probability mass quickly toward repair-friendly tokens in both stages. This leads to stronger advantages and faster learning progress. Second, F-RL begins from a pretrained checkpoint that, while syntactically accurate, is less flexible. The supervised prior constrains exploration, resisting large probability shifts toward the exact token substitutions favored by TSSR. Consequently, F-RL converges faster but achieves smaller absolute gains in validity and novelty.

\begin{table}[!t]
  \footnotesize
  \caption{Reinforcement-learning hyperparameters shared across P-RL and F-RL.}
  \label{tab:rl-hparams}
  \centering
  \begin{tabular}{@{}lcc@{}}
    \toprule
    \textbf{Hyperparameter} & \textbf{Pure RL} & \textbf{Fine-Tuning RL} \\
    \midrule
    Rollout steps (\texttt{steps\_per\_epoch})  & 512            & 512 \\
    Epochs                                      & 1{,}000        & 1{,}000 \\
    Batch size                                  & 512            & 512 \\
    PPO updates per collect                     & 1              & 1 \\
    Learning rate (\(\eta\))                    & \(1\times10^{-4}\) & \(1\times10^{-8}\) \\
    Discount factor (\(\gamma\))                & 0.99           & 0.99 \\
    GAE parameter (\(\lambda\))                 & 0.95           & 0.95 \\
    PPO clip coefficient (\(\varepsilon\))      & 0.20           & 0.20 \\
    Entropy coefficient (\(c_s\))               & 0.01           & 0.01 \\
    Value-function coefficient (\(c_v\))        & 0.50           & 0.50 \\
    Max gradient norm                           & 0.50           & 0.50 \\
    \bottomrule
  \end{tabular}
  \vspace{1cm}
\end{table}

Throughput differences follow from computational cost rather than model quality. Policies that perform and accept more repair attempts invoke RDKit more frequently, reducing overall throughput but often achieving higher final reward and validity especially visible in P-RL. In practical deployments, compute efficiency (proposals per second) and learning efficiency (reward improvement per update) should be tuned independently: the former depends on proposal depth and diagnostic evaluation, while the latter is governed by optimizer settings, entropy regularization, and PPO update aggressiveness.

Across P-RL runs, we observe a consistent link between “edit effort" and molecular validity. When the swap efficiency term \(f_{\mathrm{swap}}\) increases and the chemical error term \(f_{\mathrm{err}}\) increases, validity scores rise proportionally, confirming that the reward functions as a soft distance to feasibility. Local edits that make a molecule appear more chemically plausible are reinforced through PPO, gradually biasing the token distribution toward feasible structures. Because the reward is decomposed into interpretable components, progress can be easily audited at the token and substructure level. This transparency allows failure modes such as inconsistent ring indices, unbalanced branches, or bracket mismatches to be identified and corrected systematically. As the policy improves, an increasing fraction of previously invalid molecules becomes repairable: in the P-RL setting, the average number of molecules that can be repaired from syntactically invalid to chemically valid increases from \(478\) to \(882\). By our metrics, TSSR therefore improves both the quality and fixability of invalid outputs, which, under the interpretation of invalid SMILES proposed by Skinnider et al \cite{Skinnider2024InvalidSMILES}, may indicate a corresponding increase in the quality of the valid molecules generated.

Several limitations remain, both general to reinforcement learning and specific to molecular generation. Result variability across random seeds is most pronounced in F-RL due to stronger exploration. Stabilizing mechanisms such as advantage normalization, mild entropy annealing, adaptive learning rates based on KL divergence, or explicit KL targets in PPO could reduce variance without restricting exploration. Another limitation is length bias: shorter SMILES are easier to repair and thus tend to earn higher rewards. Incorporating length aware scoring, weak priors centered on the training distribution, or soft penalties on excessively short molecules could alleviate this bias once syntax reliability is established. Despite these challenges, TSSR demonstrates a key strength: it can teach a model to generate syntactically and chemically valid molecules from scratch using only token-frequency priors and RDKit diagnostics. This is achieved with modest resources a single environment, a lightweight GRU policy, and roughly \(9{,}000\) gradient steps.

The Two Stage Swap Reward (TSSR) provides a compact, interpretable mechanism for converting terminal molecular objectives into informative token level signals. It improves validity and novelty and integrates seamlessly with standard reinforcement learning frameworks. With richer objectives, larger token sets, and scaled training, TSSR style rewards are well positioned to become practical building blocks for de novo molecular design pipelines in computational chemistry and bioinformatics.

\section{Methods}\label{sec11}
\begin{figure}[!htbp]
    \begin{adjustbox}{center}
        \includegraphics[scale=0.85]{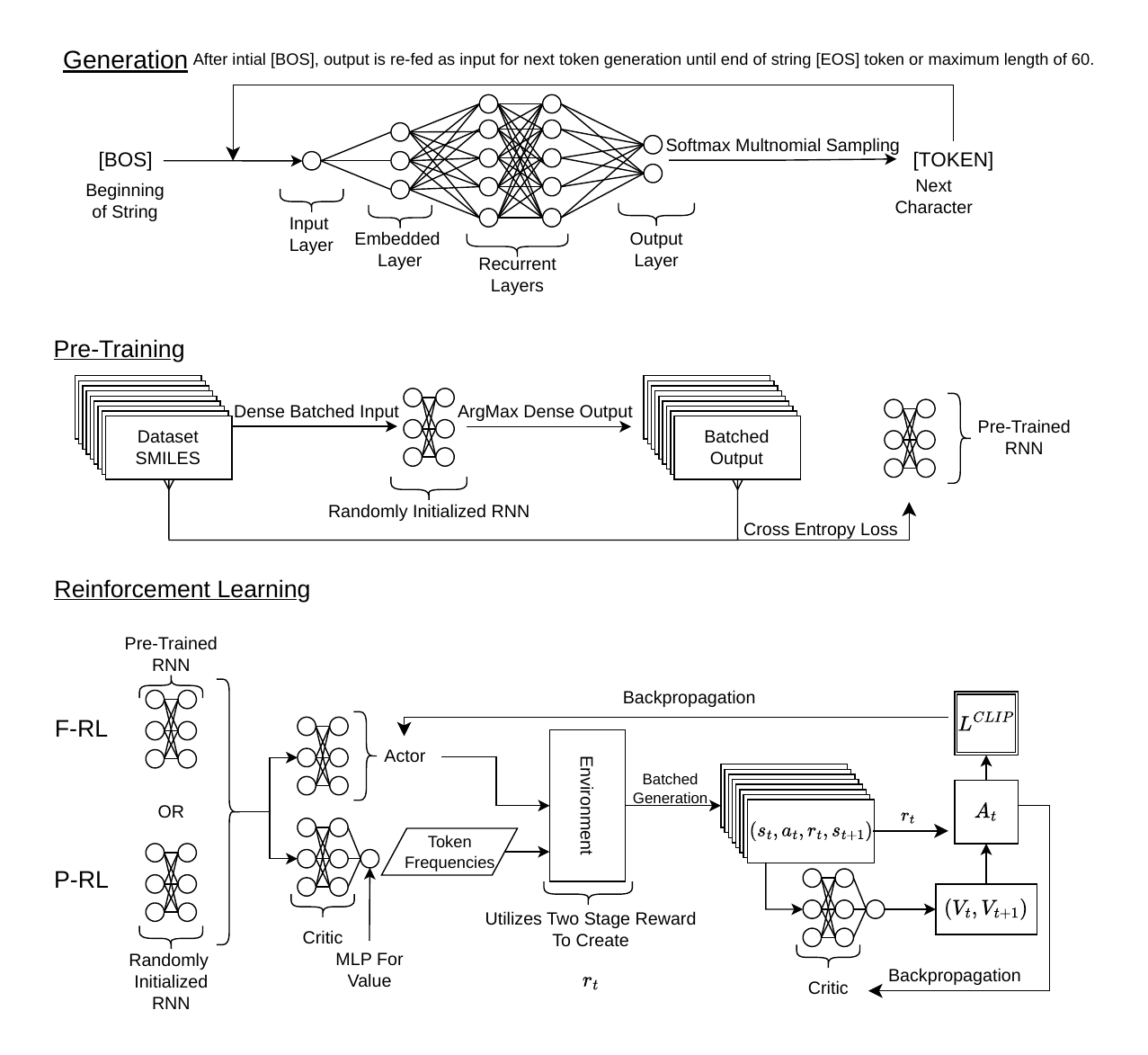}
    \end{adjustbox}
    \caption{TSSR workflow: Generation, Pre-Training, Reinforcement Learning}
    \label{fig:methods_figure}
\end{figure}

An overview of the TSSR workflow and model architecture are illustrated in \textbf{Figure~\ref{fig:methods_figure}}. 

\subsection{Reinforcement Learning Formulation}
\label{sec:generation-rl}

Molecular generation using TSSR is formulated as a finite horizon Markov Decision Process (MDP). At each step $t$, the state represents the partial SMILES sequence $s_{1:t-1}$, and the action $a_t \in \mathcal{V}$ appends one token from the vocabulary $\mathcal{V}$. The episode terminates when the end of sequence token (\texttt{[EOS]}) is produced or when the maximum sequence length $T_{\max} = 60$ is reached. Intermediate rewards are set to zero ($r_t = 0$ for $t < T$), and the terminal reward $r_T = R(s)$ is determined by the TSSR value. The discounted return is computed as

\begin{equation}
    G_t = \sum_{k=t}^{T} \gamma^{k-t} r_k,    
\end{equation}

where the discount factor is $\gamma = 0.99$. Sequences that do not contain the \texttt{[EOS]} token are truncated at $T_{\max}$ and evaluated using the same terminal reward. Each token index in $[0, |\mathcal{V}| - 1]$ is mapped to a continuous embedding through a lookup table (Dictionary), forming the input representation for the policy network.

Training is conducted using Proximal Policy Optimization (PPO) with a single vectorized environment ($n_{\text{env}} = 1$). Two training configurations are explored under the same reward formulation. In the P-RL configuration, a GRU based policy is trained from scratch using only TSSR feedback. In the F-RL configuration, the same GRU model is initialized from a pretrained SMILES language model and then optimized using TSSR. Comparing these configurations allows assessment of how chemically informed reward shaping interacts with policy initialization. The results show that pretrained representations improve convergence speed, enhance molecular validity, and increase training stability during reinforcement optimization.

\subsection{Model Architecture}
\label{sec:model-architecture}

The SMILES generator is implemented as a recurrent neural network (RNN) based on a gated recurrent unit (GRU) architecture. The network receives tokenized SMILES strings and incrementally constructs molecular sequences one token at a time. The RNN consists of an embedding layer of dimension $2|\mathcal{V}|$, followed by dropout regularization with rate $p = 0.2$, a three layer GRU with 512 hidden units per layer (also with dropout $p = 0.2$), and a linear projection head that maps the concatenation of the GRU output and hidden state to a $|\mathcal{V}|$-way softmax distribution over possible next tokens. Input–hidden weights are initialized using Xavier uniform initialization, while hidden–hidden weights are initialized orthogonally to maintain stable gradients.

At each time step $t$, the GRU maintains a hidden state $h_t \in \mathbb{R}^{d_h}$ that encodes information from previously generated tokens. This hidden state is updated through a series of gating operations that control how new and past information are combined. The reset gate $r_t$ determines how much of the previous state is ignored, the update gate $z_t$ decides how much new information is incorporated, and the candidate activation $n_t$ represents a filtered combination of the input and past context. The resulting hidden state $h_t$ balances both memory retention and adaptation to new inputs, allowing the model to capture long term dependencies without instability.

During generation, the initial hidden state is set to zero ($h_0 = \mathbf{0}$). At each step, the current token is embedded, passed through the GRU, projected into a vector of logits, and sampled from the softmax distribution to produce the next token. This process repeats until the end of sequence token (\texttt{[EOS]}) is emitted, forming a complete SMILES strings. The generation process is illustrated in Figure~\ref{fig:methods_figure}.

\paragraph{Pre-Training procedure}
For pre-training, the GRU model is pretrained for one epoch on the MOSES dataset using mixed precision (fp16) to improve computational efficiency and numerical stability. Optimization is performed with the Adam~\cite{kingma2017adammethodstochasticoptimization} optimizer ($\beta_1 = 0.9$, $\beta_2 = 0.999$) and gradient norm clipping at 10 to prevent gradient explosion, a common issue in recurrent models. The learning rate is selected automatically using Lightning’s range test across $[10^{-6}, 1]$ on a logarithmic scale, with the optimal value corresponding to the steepest loss descent.

To pre-process the dataset, each SMILES sequence is padded to a fixed length of 60 tokens, including the \texttt{[BOS]} and \texttt{[EOS]} markers. During pre-training A binary mask $m_t \in \{0,1\}$ is applied to exclude padding tokens (\texttt{[PAD]}) from the loss calculation. The model is trained to predict the next token in a sequence using a masked cross entropy loss with an additional confidence penalty that discourages overconfident predictions. Let $z_t$ denote the unnormalized logits at time step $t$, $p_t = \mathrm{softmax}(z_t)$ the predicted probability distribution over the vocabulary $\mathcal{V}$, and $y_t$ the target token. The training objective is defined as
\begin{equation}
\label{eq:conf-penalty}
\mathcal{L}
= \frac{1}{\sum_t m_t}\sum_{t=1}^{T}
m_t\!\Big[-\log p_t(y_t) + \beta \sum_{i\in\mathcal V} p_t(i)\,\log p_t(i)\Big],
\end{equation}
where $\beta = 0.1$ controls the strength of the confidence penalty.

The first term in Eq.~\eqref{eq:conf-penalty} represents the standard negative log likelihood that encourages the model to assign high probability to the correct next token. The second term acts as an entropy regularizer that penalizes overly confident predictions by introducing a small positive penalty when the probability distribution becomes too sharp. This combination balances accuracy and generalization, mitigating overfitting to frequent molecular substructures while maintaining diversity in generated SMILES sequences. Through this pre-training phase, the GRU learns chemically and syntactically meaningful representations that serve as a strong initialization for reinforcement learning under the TSSR framework.

\paragraph{Metrics and Evaluation}
\label{Metrics and Evaluation}
To evaluate model performance, 10{,}000 SMILES strings are generated then evaluated across a range of chemical and structural metrics. These measures capture not only the syntactic and chemical correctness of generated molecules but also their diversity, novelty, and drug-likeness.

\textbf{Drug likeness and synthesis quality.}
The practical quality of generated molecules are evaulted using two established metrics. The Quantitative Estimate of Drug likeness (QED) score, measures pharmacological realism on a scale from 0 to 1. The Synthetic Accessibility (SA) score, estimates how easy a molecule would be to synthesize in practice. Their means, medians, and the fractions of molecules are reported, satisfying typical cutoffs (\(QED \ge 0.6\), \(SA \le 3\), and \(SA \le 5\)).

\textbf{Diversity.}
To understand how broadly the model explores chemical space, two complementary measures of diversity are computed.  
The first is nearest neighbor diversity, defined as the mean Tanimoto distance between each molecule’s ECFP4 fingerprint (radius 2, 2048 bits) and its most similar neighbor within the generated set. A higher value indicates greater chemical variety.  
The second is scaffold diversity, calculated as the number and fraction of unique Bemis-Murcko scaffolds among valid molecules. This reflects structural diversity at the core framework level.

\textbf{Structural similarity.}
Finally, to compare the overall structure of generated molecules to the reference set, scaffold similarity is computed using the average maximum Tanimoto similarity between Bemis-Murcko scaffolds from both sets. Lower similarity corresponds to more structurally novel generations.

\subsection{Two Stage Swap Reward}
\label{ssec:reward}

To provide structured and interpretable reinforcement signals, the TSSR framework is introduced as the core of the training strategy. Instead of assigning a binary validity score to generated SMILES sequences, TSSR delivers graded feedback that measures both syntactic correctness and chemical plausibility. The method operates in two sequential stages syntax repair and chemical refinement each contributing complementary components to the overall reward. The framework is dataset agnostic and relies only on token frequency statistics and RDKit chemistry diagnostics~\cite{greg_landrum}.

In the first stage, the algorithm attempts to repair syntactically invalid SMILES strings through minimal local substitutions. Given a sequence $s = (c_1, \ldots, c_T)$ over the vocabulary $\mathcal{V}$, token positions are randomly shuffled, and for each position $i$, a small set of alternative tokens $\mathcal{K}$ is sampled from $\mathcal{V}$ without replacement, weighted by empirical token frequencies. Each token candidate $c'$ replaces $s[i]$ to form a new sequence $s'$, which is parsed using \texttt{MolFromSmiles}$(s', \texttt{sanitize{=}}\texttt{False})$. If parsing succeeds, $s'$ is returned as a repaired molecule. Otherwise, the failure counter $N_{\mathrm{fail}}$ is incremented. The search terminates once a valid substitution is found or all positions and candidates have been exhausted. The swap efficiency term 
\begin{equation}
    f_{\mathrm{swap}} = \frac{1}{1 + N_{\mathrm{fail}}}
\end{equation}

quantifies how many unsuccessful repair attempts ($N_{\mathrm{fail}}$) were required before achieving valid syntax. Higher values of $f_{\mathrm{swap}}$ therefore indicate that fewer substitutions were needed, rewarding efficient corrections that achieve syntactic validity with minimal effort. This repair process is formalized in Algorithm~\ref{alg:syntax}, which implements the syntax fixing loop described above.

\begin{algorithm}[!t]
\footnotesize
  \caption{Syntax Repair}
  \label{alg:syntax}
  \begin{algorithmic}[1]
    \Procedure{TrySyntaxFix}{$s$}
      \State $C \gets \textsc{Shuffle}(\{1,\dots,|s|\})$ \Comment{random position order}
      \ForAll{$i \in C$}
        \State $\mathcal K \gets \textsc{SampleNoDup}(\mathcal V,k_{\mathrm{subst}};\text{prob}=\texttt{token\_freq})$
        \ForAll{$c' \in \mathcal K$}
          \If{$c' = s[i]$} \textbf{continue} \EndIf
          \State $s' \gets s$; \quad $s'[i]\gets c'$
          \State $m' \gets \MolFromSmiles(s',\texttt{sanitize}{=}\texttt{False})$
          \If{$m'\neq\texttt{None}$}
            \State \Return $s'$ \Comment{first successful repair}
          \EndIf
        \EndFor
      \EndFor
      \State \Return \texttt{None} \Comment{repair failed}
    \EndProcedure
  \end{algorithmic}
\end{algorithm}

Once a syntactically valid sequence is obtained either directly from the model or through repair the second stage focuses on improving chemical plausibility by reducing RDKit detected errors such as valence violations, aromaticity inconsistencies, or incorrect atom connectivity. The molecule is sanitized, and the number of detected issues $|E(m)|$ is recorded. If no errors are found ($|E(m)|=0$), the molecule is chemically valid and receives the highest reward. Otherwise, token positions are again shuffled, and for each index $i$, candidate replacements are sampled and tested as in Stage One. Substitutions that yield fewer chemistry problems are accepted as new best solutions, and the process repeats until no further improvement is achieved or the molecule becomes fully valid. This iterative refinement procedure is detailed in Algorithm~\ref{alg:chem}, which provides the implementation corresponding to the chemical-error reduction process described here. The outputs of this stage define two diagnostic measures:

\begin{align}
    f_{\mathrm{err}} = \frac{|E_0| - |E^*|}{\max(1, |E_0|)}, & \qquad
    f_{\mathrm{dist}} = 1 - \frac{|E^*|}{E_{\max}}.
\end{align}

Here, $E_0$ denotes the set of chemical errors detected in the initial molecule, $E^*$ is the smallest (best) error set found after all attempted substitutions, and $E_{\max} = 12$ defines the maximum number of error categories considered by the diagnostics (e.g., valence, aromaticity, charge balance). The term $f_{\mathrm{err}}$ measures the fractional reduction in chemical problems, taking values in $[-1,1]$ where larger values indicate greater improvement, while $f_{\mathrm{dist}}$ provides a normalized measure of how chemically sound the final structure is relative to the diagnostic scale, with $f_{\mathrm{dist}}=1$ corresponding to full validity.

\begin{algorithm}[!t]
\footnotesize
  \caption{Chemical Error Reduction}
  \label{alg:chem}
  \begin{algorithmic}[1]
    \Procedure{TryReduceChemProblems}{$s$}
      \State $m^* \gets \MolFromSmiles(s,\texttt{sanitize}{=}\texttt{True})$
      \State $e^* \gets |E(m^*)|$, \quad $e_0\gets e^*$, \quad $N_{\text{fail}}\gets0$
      \If{$e^*=0$} \State \Return $(0,\,0,\,1)$ \EndIf
      \State $m \gets \MolFromSmiles(s,\texttt{sanitize}{=}\texttt{False})$
      \ForAll{$i \in \textsc{Shuffle}(\{1,\dots,|s|\})$}
        \State $\mathcal K\gets\textsc{SampleNoDup}(\mathcal V,k_{\mathrm{subst}};\text{prob}=\texttt{token\_freq})$
        \ForAll{$c'\in\mathcal K$}
          \If{$c'=s[i]$} \textbf{continue} \EndIf
          \State $s' \gets s$; \quad $s'[i]\gets c'$
          \State $m' \gets \MolFromSmiles(s',\texttt{sanitize}{=}\texttt{False})$
          \If{$m'=\texttt{None}$} \State $N_{\text{fail}}\mathrel{{+}{=}}1$; \textbf{continue} \EndIf
          \State $e\gets |E(m')|$
          \If{$e<e^*$}
            \State $(m^*,e^*)\gets(m',e)$; \quad $s\gets s'$
            \If{$e^*=0$}
              \State \Return $\bigl(N_{\text{fail}},\frac{e_0-e^*}{\max(1,e_0)},1-\tfrac{e^*}{E_{\max}}\bigr)$
            \EndIf
            \State \textbf{break} \Comment{accept improvement and restart outer loop}
          \Else
            \State $N_{\text{fail}}\mathrel{{+}{=}}1$
          \EndIf
        \EndFor
      \EndFor
      \State \Return $\bigl(N_{\text{fail}},\frac{e_0-e^*}{\max(1,e_0)},1-\tfrac{e^*}{E_{\max}}\bigr)$
    \EndProcedure
  \end{algorithmic}
\end{algorithm}

The overall reward integrates all three components using nonnegative weights $\lambda_{\mathrm{swap}}$, $\lambda_{\mathrm{err}}$, and $\lambda_{\mathrm{dist}}$, which satisfy $\lambda_{\mathrm{swap}}+\lambda_{\mathrm{err}}+\lambda_{\mathrm{dist}}=1$:
\begin{equation}
    \label{eq:reward}
R(s) =
\begin{cases}
  \lambda_{\mathrm{swap}}f_\text{swap} + \lambda_{\mathrm{err}}f_\text{err} + \lambda_{\mathrm{dist}}f_\text{dist}, & \text{if $s$ is syntactically valid}, \\[3pt]
  -\tfrac{1}{2} + \lambda_{\mathrm{swap}}f_\text{swap} + \lambda_{\mathrm{err}}f_\text{err} + \lambda_{\mathrm{dist}}f_\text{dist}, & \text{if $s$ is repaired from invalid}, \\[3pt]
  -1, & \text{if $s$ remains invalid after repair.}
\end{cases}
\end{equation}

In Eq.~\eqref{eq:reward}, the term $R(s)$ denotes the final molecular reward. The weights $\lambda_{\mathrm{swap}}, \lambda_{\mathrm{err}},$ and $\lambda_{\mathrm{dist}}$ control the relative influence of syntactic efficiency, chemical improvement, and chemical soundness, respectively. The partial penalty of $-\tfrac{1}{2}$ rewards molecules that recover validity while acknowledging the initial parsing failure, whereas completely invalid sequences incur the full penalty of $-1$.

This formulation provides continuous, interpretable feedback that decomposes the learning signal into syntactic and chemical components. Molecules that are both syntactically valid and chemically sound obtain the highest rewards, partially repaired molecules receive proportional credit, and unrepaired strings are penalized but still provide gradient information to guide learning. The smooth reward surface stabilizes reinforcement learning and enables gradual, interpretable improvement in both grammar and chemical validity. Each stage evaluates at most $T_{\max}k_{\mathrm{subst}}$ substitution candidates, where $T_{\max}$ is the maximum sequence length and $k_{\mathrm{subst}}$ is the number of token proposals per position. On decent GPU hardware, TSSR achieves throughput between $10^{5}$ and $10^{5}$ tokenevaluations per second, enabling efficient integration into large scale molecular generation workflows.

\subsection{Training Structure}\label{sec:train-structure}

To evaluate the effectiveness of TSSR in different learning regimes, two reinforcement learning configurations are considered which share the same reward formulation and training protocol but differ in network initialization. In the first setting, referred to as P-RL, the GRU policy network is initialized with random weights and trained from scratch using Proximal Policy Optimization (PPO). The network functions as the actor, while the critic network replicates the actor’s GRU backbone and appends a lightweight multilayer perceptron head that outputs a scalar value estimate. Both actor and critic are optimized jointly for 1{,}000 epochs with a batch size of 512 in a single vectorized environment using the TSSR reward.

In the second setting, F-RL, the same GRU architecture is first pretrained for one epoch on the MOSES dataset to learn general SMILES syntax and token dependencies. The pretrained weights are then transferred to initialize the PPO actor, while the critic again mirrors the GRU backbone with an additional value head. Training proceeds for 1{,}000 epochs with identical batch size and environment configuration as in P-RL, enabling a controlled comparison between models trained from random initialization and those initialized with learned chemical priors. This setup allows us to assess how prior sequence knowledge influences convergence speed, stability, and the generation of syntactically and chemically valid molecules under the TSSR reward.

\subsection{Policy optimization for terminal reward SMILES generation}

SMILES generation is modeled as a sequential decision-making process in which the agent constructs a molecule one token at a time. At each step \(t\), the \emph{state} \(s_t\) is defined as the sequence of all tokens generated so far, beginning from the start-of-sequence token \texttt{[BOS]}. The agent then selects an \emph{action} \(a_t\), which corresponds to choosing the next token from the vocabulary \(\mathcal{V}\). The policy \(\pi_\theta(a_t \mid s_t)\) parameterized by \(\theta\) represents the probability distribution over possible next tokens given the current prefix \(s_t\). This distribution determines which character (atom, bond, or symbol) is appended to the partial SMILES. The process continues until the agent outputs the end of sequence token \texttt{[EOS]} or the predefined maximum sequence length of 60 tokens (including \texttt{[BOS]} and \texttt{[EOS]}) is reached. The objective of learning is to adjust the policy parameters \(\theta\) so that the expected reward \(R(s)\) for complete sequences \(s = (c_1, \dots, c_T)\) is maximized. Here, \(R(s)\) denotes the molecular reward calculated using the TSSR framework described earlier (Eq.~\eqref{eq:reward}).

Since both syntactic validity and chemical diagnostics are only meaningful for a fully formed molecule, the reward is provided exclusively at the terminal step:
\begin{equation}
\label{eq:terminal-reward}
r_t =
\begin{cases}
0, & t<T-1,\\
R(s), & t=T-1,
\end{cases}
\qquad
V(s_T)=0.
\end{equation}
In Eq.~\eqref{eq:terminal-reward}, the intermediate rewards \(r_t\) are set to zero for all steps before the last token (\(t<T-1\)), because partial SMILES strings cannot be meaningfully evaluated for chemical correctness. The terminal reward \(R(s)\) is assigned when the final token is generated (\(t=T-1\)), and the value function at the terminal state \(V(s_T)\) is defined as zero to mark the end of the trajectory. This convention allows the critic to bootstrap properly from the terminal state when estimating future returns.

The total discounted return observed from step \(t\) is then given by
\begin{equation}
\label{eq:returns}
G_t = \gamma^{\,T-1-t}\,R(s)\quad (t<T-1),
\qquad
G_{T-1} = R(s),
\end{equation}
where \(G_t\) represents the expected cumulative reward from time step \(t\) onward, and \(\gamma \in (0,1]\) is the discount factor that controls the relative weighting of immediate versus distant rewards. The term \(\gamma^{T-1-t}\) scales the terminal reward such that decisions made earlier in the sequence receive a  smaller share of the final feedback, preventing the early tokens from dominating the gradient updates. For the terminal step (\(t = T-1\)), the return simply equals the actual molecular reward \(R(s)\).

To enable stepwise learning, the temporal difference (TD) residuals are computed which quantify how well the critic’s value estimates align with the observed outcomes:
\begin{align}
\label{eq:td-residuals}
\delta_t &= \gamma\,V_{\theta_{\text{old}}}(s_{t+1}) - V_{\theta_{\text{old}}}(s_t), && t<T-1,\\
\delta_{T-1} &= R(s) - V_{\theta_{\text{old}}}(s_{T-1}).
\end{align}
In these expressions, \(V_{\theta_{\text{old}}}(s_t)\) is the critic’s estimated value for the current state \(s_t\) under the behavior policy with parameters \(\theta_{\text{old}}\). For intermediate steps (\(t<T-1\)), the TD residual \(\delta_t\) measures the difference between the discounted next state value \(\gamma V_{\theta_{\text{old}}}(s_{t+1})\) and the current prediction \(V_{\theta_{\text{old}}}(s_t)\), indicating how much the critic underestimated or overestimated future rewards. At the final step, \(\delta_{T-1}\) directly compares the observed molecular reward \(R(s)\) with the critic’s final value estimate, ensuring that the value function aligns with true molecular performance.

To propagate the terminal reward across the entire sequence, Generalized Advantage Estimation (GAE) is employed:
\begin{equation}
\label{eq:gae}
\hat A_t = \sum_{l=0}^{T-1-t} (\gamma\lambda)^l \,\delta_{t+l}.
\end{equation}
Here, \(\hat A_t\) denotes the advantage estimate for step \(t\), which represents how much better the chosen action \(a_t\) was than the critic’s expectation at that state. The factor \(\lambda \in [0,1]\) determines the temporal smoothing of credit assignment. When \(\lambda\) is large (close to 1), the advantage incorporates information from many future steps, distributing the terminal reward more evenly across earlier tokens and when \(\lambda\) is small, the advantage relies primarily on immediate TD signals, reducing variance but shortening the effective credit horizon. In molecular generation, a larger \(\lambda\) ensures that beneficial early choices such as correct ring numbering or bracket balancing receive positive feedback even before the molecule is complete.

The critic’s learning target is computed as
\begin{equation}
    \hat G_t = \hat A_t + V_{\theta_{\text{old}}}(s_t),
\end{equation}

where \(\hat G_t\) acts as a fixed regression target during the policy update. The critic is trained to minimize the squared error between its current prediction \(V_\theta(s_t)\) and this target, aligning the estimated values with the observed returns from the behavior policy.

To stabilize policy updates, the clipped Proximal Policy Optimization (PPO) objective is adopted. Let
\begin{equation}
    r_t(\theta)=\frac{\pi_\theta(a_t\mid s_t)}{\pi_{\theta_{\text{old}}}(a_t\mid s_t)}
\end{equation}

be the likelihood ratio between the new policy and the previous one for the sampled action \(a_t\). The PPO clipped surrogate loss is defined as
\begin{equation}
\label{eq:ppo-clip}
\mathcal{L}^{\mathrm{CLIP}}(\theta)
= \hat{\mathbb{E}}_t\!\left[
\min\!\big(
r_t(\theta)\,\hat A_t,\;
\mathrm{clip}(r_t(\theta),\,1-\varepsilon,\,1+\varepsilon)\,\hat A_t
\big)
\right],
\end{equation}
where \(\varepsilon=0.2\) is the clipping threshold. The ratio \(r_t(\theta)\) quantifies how much the probability of the chosen action changes after the policy update: if \(\hat A_t>0\), increasing that probability improves performance, while if \(\hat A_t<0\), reducing it is beneficial. The clipping function \(\mathrm{clip}(r_t(\theta),1-\varepsilon,1+\varepsilon)\) restricts the ratio to a bounded interval to prevent excessively large policy updates that could destabilize learning or collapse sequence diversity.

The value function is trained with PPO-v1 style value clipping to further limit large changes in value estimates:
\begin{equation}
\label{eq:value-clipping}
\begin{aligned}
L_v(t) = \max\Bigl(& (V_\theta(s_t)-\hat G_t)^2,\,
\bigl(\operatorname{clip}(V_\theta(s_t),\,
      V_{\theta_{\text{old}}}(s_t)-\epsilon_v,\,
      V_{\theta_{\text{old}}}(s_t)+\epsilon_v) - \hat G_t \bigr)^2
\Bigr),
\end{aligned}
\end{equation}
where \(\epsilon_v\) defines the clipping range for value changes. The first term penalizes deviations between the new value estimate and the target, while the second term ensures that the critic does not overreact to noise by restricting value shifts to a small neighborhood of the previous estimate.

The complete PPO objective combines the policy, value, and entropy components:
\begin{equation}
\label{eq:ppo-full}
\begin{aligned}
\mathcal{L}^{\mathrm{PPO}}(\theta) =\;&
-\,\hat{\mathbb{E}}_t\!\big[\mathcal{L}^{\mathrm{CLIP}}(\theta)\big]
+ c_v\,\hat{\mathbb{E}}_t\!\big[L_v(t)\big]
- c_s\,\hat{\mathbb{E}}_t\!\big[\mathcal H(\pi_\theta(\cdot\!\mid\!s_t))\big],
\end{aligned}
\end{equation}
where \(c_v = 0.5\) and \(c_s = 0.01\) are coefficients balancing the critic loss and entropy regularization terms. The final term, \(\mathcal H(\pi_\theta(\cdot\!\mid\!s_t))\), represents the entropy of the token probability distribution, which encourages exploration by preventing the policy from becoming overly deterministic. This regularization is especially important in SMILES generation: during early training it promotes exploration around near valid sequences, and later it allows the model to consider alternative chemically meaningful corrections instead of repeating a single preferred pattern.

Together, these formulations transform the sparse, terminal molecular reward into dense, stepwise learning signals that effectively guide sequence generation. 
Early in training, the policy learns to form syntactically valid structures that are close to parseable SMILES.
As training progresses, it increasingly focuses on chemically consistent modifications that reduce RDKit detected errors. 
Each mathematical component from the TD residuals in Eq.~\eqref{eq:td-residuals} to the entropy term in Eq.~\eqref{eq:ppo-full} contributes to stabilizing learning and improving both syntactic and chemical validity of the generated molecules under the TSSR reward framework.

\subsection{Training setup, environment, and hyperparameters}
\label{ssec:training}

The TSSR framework was implemented using PyTorch Lightning and trained on an NVIDIA RTX 3090 GPU equipped with 24 GB of GDDR6X memory. The policy is trained using short, regular rollouts followed by a single conservative update per data collection. Each training epoch gathers \texttt{steps\_per\_epoch}$=512$ token level transitions, organized into smaller micro-collections of \texttt{steps\_per\_collect}$=60$, resulting in approximately nine collections per epoch. After each collect, one Proximal Policy Optimization (PPO) update is performed (\texttt{repeat\_per\_collect}$=1$) using a batch size of \texttt{batch\_size}$=512$. Since all collected transitions fit into a single batch, no additional mini-batching is required, and effectively one PPO epoch is executed per batch. Increasing \texttt{repeat\_per\_collect} to 2–4 reduced sample requirements but consistently degraded chemical quality and novelty, so it is fixed at one update per collect throughout training.

The reinforcement-learning environment follows the Gymnasium~0.28.1~\cite{towers2024gymnasium} interface with discrete observation and action spaces of size $|\mathcal{V}|$, where $\mathcal{V}$ is the SMILES token vocabulary. At each step $t$, the state $s_t$ corresponds to the partial SMILES strings, and the action $a_t$ is the next token sampled from $\mathcal{V}$. Episodes terminate either when the end of sequence token \texttt{[EOS]} is emitted or when the length cap $T_{\max}=60$ is reached (counting \texttt{[BOS]} and \texttt{[EOS]}). Intermediate rewards are zero ($r_t=0$ for $t<T-1$), and the terminal molecular reward $R(s)$, defined in Eq.~\eqref{eq:reward}, is returned only upon termination. If the generated sequence ends with \texttt{[EOS]}, that token is removed before computing $R(s)$ to ensure consistency.

Several hyper-parameters determine the learning stability and performance of PPO.  
The clipping coefficient $\varepsilon$ bounds how much the token probabilities may change in one policy update. Larger $\varepsilon$ allows faster adaptation but can destabilize learned grammar, whereas smaller $\varepsilon$ slows convergence. A value of $\varepsilon=0.2$ provides a robust compromise for terminal only rewards.  
The critic weight $c_v$ controls the relative influence of value function learning in the overall PPO loss. The value of $c_v=0.5$ balances critic accuracy and policy improvement.  
The entropy coefficient $c_s=0.01$ maintains exploration by discouraging overly peaked token distributions, which is especially beneficial during Stage One syntax repair and Stage Two chemical refinement when the model must test alternative valid sequences.  
The Generalized Advantage Estimation parameter $\lambda=0.95$ determines how far the terminal reward propagates backward through the sequence, and the discount factor $\gamma=0.99$  down weights very early actions while preserving nearly all of the terminal signal.  
Learning rate $\eta$ is tied to initialization: in P-RL, the policy starts from random weights and uses $\eta=10^{-4}$ for faster convergence while in F-RL, the policy begins from a pretrained SMILES language model and adopts $\eta=10^{-8}$ to avoid overwriting syntactic knowledge.  
Finally, the global gradient norm is clipped at 0.5 to prevent rare large updates from destabilizing training.  
All of these settings are summarized in Table~\ref{tab:rl-hparams}, where the parameters are grouped into rollout scale (number of steps and epochs), optimization settings (learning rate, clipping), and regularization coefficients ($c_v$, $c_s$, and gradient norm).

Short rollouts of 512 token steps combined with one PPO update per collect keep the data on policy and minimize stale trajectories. Larger numbers of PPO repeats per collect would improve sample efficiency but were observed to induce on policy drift, lowering molecular validity and novelty. Therefore, stability is prioritized over sample reuse in all reported experiments.

TSSR introduces additional parameters that govern substitution search and the weighting of reward components. The substitution budget $k_{\mathrm{subst}}=8$ defines how many alternative tokens are sampled per position in both syntax repair and chemical error reduction. This budget provides enough diversity to discover viable repairs without converting Stage One into a broad random search.  
The reward weights $\lambda_{\mathrm{swap}}$, $\lambda_{\mathrm{err}}$, and $\lambda_{\mathrm{dist}}$ control the relative contribution of the three feedback components from Eq.~\eqref{eq:reward}. Specifically, $\lambda_{\mathrm{err}}=0.50$ emphasizes reduction of RDKit detected chemistry problems, $\lambda_{\mathrm{dist}}=0.30$ rewards proximity to a fully valid molecule, and $\lambda_{\mathrm{swap}}=0.20$ rewards efficient syntax repair with fewer failed substitutions. These coefficients sum to one and maintain a balanced objective between chemical improvement and syntactic efficiency. Table~ \ref{tab:swap-hparams} lists these reward specific parameters for both P-RL and F-RL.

\begin{table}[!t]
  \footnotesize
  \caption{Two Stage Swap Reward hyperparameters controlling substitution budget and reward weighting.}
  \label{tab:swap-hparams}
  \centering
  \begin{tabular}{@{}lcc@{}}
    \toprule
    \textbf{Hyperparameter} & \textbf{Pure RL} & \textbf{Fine-Tuning RL} \\
    \midrule
    Substitution budget (\(k_{\mathrm{subst}}\)) & 8 & 8 \\
    Syntax-efficiency weight (\(\lambda_{\mathrm{swap}}\)) & 0.20 & 0.20 \\
    Error-reduction weight (\(\lambda_{\mathrm{err}}\))    & 0.50 & 0.50 \\
    Chemical-soundness weight (\(\lambda_{\mathrm{dist}}\)) & 0.30 & 0.30 \\
    \bottomrule
  \end{tabular}
  \vspace{1cm}
\end{table}

Taken together, these configurations yield a simple and predictable training schedule: collect a small batch of recent trajectories, apply one clipped PPO update, maintain exploration through entropy regularization, and bias the terminal reward toward chemically meaningful progress.  
Table~\ref{tab:rl-hparams} provides the optimization and rollout parameters that define the reinforcement-learning environment, while Table~\ref{tab:swap-hparams} details the reward specific coefficients used by the TSSR framework.  
Together they fully specify the experimental conditions, enabling direct reproducibility and quantitative comparison between P-RL and F-RL training regimes.

\section{Conclusion}\label{sec13}
Generating novel molecules from SMILES strings is challenging because even minor token errors can invalidate a sequence or lead to chemically implausible structures. We introduced the TSSR, a lightweight and interpretable reinforcement signal designed to address both issues. In Stage One, local token substitutions repair syntactic errors then, in Stage Two, targeted edits reduce chemical inconsistencies. Together, these stages transform a sparse terminal reward into a continuous feedback signal that the policy can effectively learn from.

We trained the model under two regimes. In P-RL, the policy was learned from scratch, while in F-RL, the model was initialized from a pretrained checkpoint on MOSES. P-RL achieved the largest improvements in validity and novelty, whereas F-RL converged faster but produced smaller quality gains. In both settings, higher validity correlated with more successful syntax repairs and fewer chemistry problems, demonstrating that the reward promotes genuine molecular improvement rather than simply exploiting parser behavior.

The method is computationally efficient. It relies only on token frequency statistics and can be executed on GPU hardware. Its modular design allows straightforward integration into broader de novo design pipelines, where it can complement property specific or task driven objectives.

Our experiments also reveal several practical insights and opportunities for improvement. First, we observed a length bias: shorter sequences are easier to repair and thus tend to receive higher rewards, leading to a slight shortening of average molecule length in P-RL. This effect can be mitigated using length aware rewards, weak priors that maintain lengths near the training distribution, minimum length constraints, or mild penalties on overly short outputs once syntax stabilization has been achieved.

Second, the current reward focuses solely on validity and general chemical plausibility. Real world molecular design requires additional objectives such as synthesizability, structural diversity, and property optimization. The TSSR framework can easily be extended with new reward terms for synthetic accessibility, scaffold novelty, and physicochemical constraints, introduced progressively as the model learns to generate valid chemistry.

Third, we restricted the token vocabulary to that of the MOSES dataset for stability and reproducibility. This simplification narrows chemical coverage. Expanding the token set to include charged atoms, metals, stereochemical markers, and frequent functional fragments would increase expressivity. Since TSSR depends only on token frequencies and RDKit diagnostics, it remains compatible with larger or customized vocabularies.

Fourth, our training setup was intentionally conservative, using a single environment and approximately \(9{,}000\) gradient steps. Many reinforcement learning systems employ multiple parallel environments and substantially longer training schedules. Further improvements are expected from vectorized environments, increased rollout counts, and more gradient updates. Incorporating KL-regularized PPO could accelerate F-RL adaptation while preserving pretrained priors.

Beyond SMILES, TSSR generalizes naturally. It can be paired with transformer based architectures, applied to DeepSMILES or SELFIES representations (Stage Two), or extended to graph based molecular generators that enforce valence and ring closure constraints directly on nodes and edges. As with any shaped reward, over optimization remains a concern: high novelty does not always imply chemical usefulness. Incorporating guardrails such as scaffold diversity metrics, property distribution matching, nearest-neighbor similarity checks, and Fréchet-like distances can ensure alignment with practical design goals.

In summary, TSSR provides a compact, interpretable, and easily deployable mechanism for teaching reinforcement learning policies to generate syntactically valid and chemically meaningful molecules. It improves validity without specialized infrastructure, integrates seamlessly with standard RL tool-chains, and offers a clear path toward richer objectives, larger vocabularies, and more scalable training setups. With these extensions, TSSR can serve as a practical foundation for future molecular design systems used by chemists and bioinformaticians alike.

\subsection{Data Availability}
The MOSES benchmark dataset used in this study can be downloaded from \url{https://github.com/molecularsets/moses}. The SMILES strings generated in this study are provided in the Supplementary data.

\subsection{Code Availability}
Source code is archived on Zenodo at \url{https://doi.org/10.5281/zenodo.17459451}, and also available at \url{https://github.com/LynaLuo-Lab/tssr-smiles-generation}.

\subsection{Acknowledgements}
This work was supported by the Cal-Bridge Program (J.E.L.), NIH grants GM130834 (Y.L.L.), and the Calpoly Pomona research fund (S.C.K).

\subsection{Author Contributions}
J.E.L. conducted research and analyzed the data. S.C.K and Y.L.L designed and supervised the project. All authors prepared the manuscript. 

\subsection{Competing Interests}
The authors declare no competing interest.
\bibliographystyle{unsrt}
\bibliography{sn-bibliography}
\end{document}